\documentclass[runningheads]{llncs}

\usepackage{graphicx}
\usepackage{comment}
\usepackage{amsmath,amssymb} % define this before the line numbering.
\usepackage{color}

\usepackage[utf8]{inputenc} % allow utf-8 input
\usepackage[T1]{fontenc}    % use 8-bit T1 fonts
\usepackage{url}            % simple URL typesetting
\usepackage{booktabs}       % professional-quality tables
\usepackage{amsfonts}       % blackboard math symbols
\usepackage{nicefrac}       % compact symbols for 1/2, etc.
\usepackage{microtype}      % microtypography
\usepackage{lipsum}
\usepackage{soul}

\usepackage{amsmath,amssymb} % define this before the line numbering.
\usepackage{subfigure}
\usepackage{hyperref}
\usepackage{cleveref}
\usepackage{multirow}
\usepackage{multicol}

\begin{document}

\pagestyle{headings}
\mainmatter
\def\ECCVSubNumber{6259}  % Insert your submission number here

\title{Orderly Disorder in Point Cloud Domain} % Replace with your title

% INITIAL SUBMISSION 
\begin{comment}
\titlerunning{ECCV-20 submission ID \ECCVSubNumber} 
\authorrunning{ECCV-20 submission ID \ECCVSubNumber} 
\author{Anonymous ECCV submission}
\institute{Paper ID \ECCVSubNumber}
\end{comment}
%******************

% CAMERA READY SUBMISSION
%\begin{comment}
\titlerunning{Orderly Disorder in Point Cloud Domain}

\author{Morteza~Ghahremani\inst{1}, Bernard~Tiddeman\inst{1}, Yonghuai~Liu\inst{2}, and Ardhendu~Behera\inst{2}}
\authorrunning{M. Ghahremani et al.}% First names are abbreviated in the running head. If there are more than two authors, 'et al.' is used.
\institute{Department of Computer Science,
  Aberystwyth University,
  Wales, UK\\
\email{\{mog9,bpt\}@aber.ac.uk}
\and Department of Computer Science,
  Edge Hill University, Lancashire, UK\\
\email{\{liuyo,beheraa\}@edgehill.ac.uk}}
%\end{comment}
%******************
\maketitle
% Papers with more than 14 pages (excluding references) will be rejected without review.
\begin{abstract}  % 70-150 words
In the real world, out-of-distribution samples, noise and distortions exist in test data. Existing deep networks developed for point cloud data analysis are prone to overfitting and a partial change in test data leads to unpredictable behaviour of the networks.
In this paper, we propose a smart yet simple deep network for analysis of 3D models using `orderly disorder' theory. Orderly disorder is a way of describing the complex structure of disorders within complex systems.  
Our method extracts the deep patterns inside a 3D object via creating a dynamic link to seek the most stable patterns and at once, throws away the unstable ones. 
Patterns are more robust to changes in data distribution, especially those that appear in the top layers.
Features are extracted via an innovative cloning decomposition technique and then linked to each other to form stable complex patterns.
Our model alleviates the vanishing-gradient problem, strengthens dynamic link propagation and substantially reduces the number of parameters.
Extensive experiments on challenging benchmark datasets verify the superiority of our light network on the segmentation and classification tasks, especially in the presence of noise wherein our network's performance drops less than 10\% while the state-of-the-art networks fail to work.
\keywords{Point cloud, deep neural network, orderly disorder, segmentation, classification}
\end{abstract}
%%%%%%%%%%%%%%%%%%%%%%%%%%%%%%%%%%%%%%%%%%%%%%%%%%%%%%%%%%%%%%
\section{Introduction}
\begin{figure*}[!ht]%##############################################
\centering
    \subfigure[]{
    \includegraphics[width=0.6\linewidth]{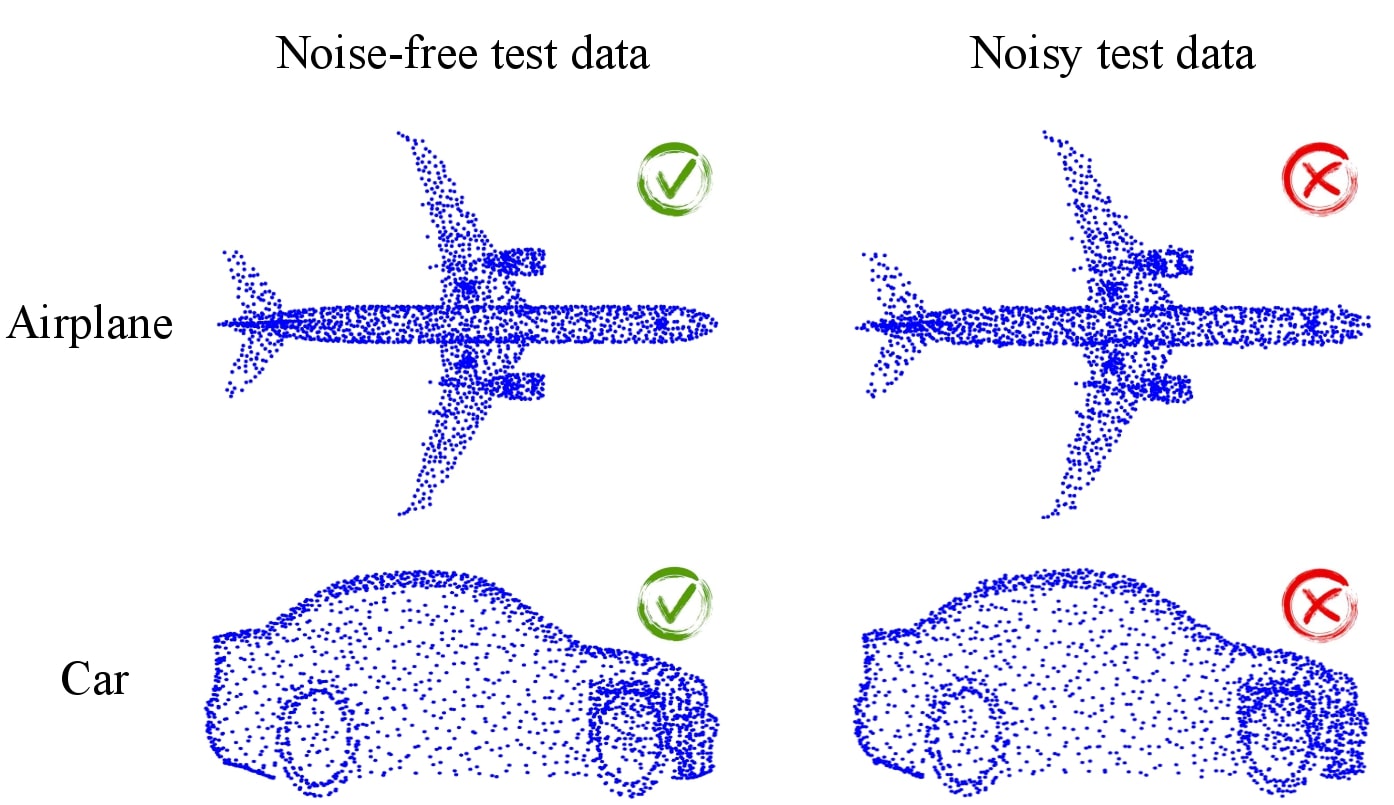}}
\caption{Classification results of most existing networks  highly depend on the distribution of training and test data. 
A partial change in the distribution of test data by adding Gaussian noise $\mathcal{N}(0,0.02)$ leads to misclassification.}
\label{fig::ircls}
\end{figure*}

Object classification and semantic segmentation of 3D models are foundations of numerous computer vision applications like autonomous driving and robot manipulation.
Thus far, a considerable number of convolutional neural  networks (CNNs) have been developed for such tasks \cite{maturana2015voxnet,wu20153d,su2015multi,qi2016volumetric,qi2017pointnet,guerrero2018pcpnet} and in most cases they yield promising results, especially when the distributions of test and train datasets are similar.
However, 3D models in the real world contain out-of-distribution samples, different samplings, noise and distortions that significantly influence their performance. Fig. \ref{fig::ircls} shows a few examples of wrong classification in the presence of noise.
%########################################

Using a large number of parameters for modeling and training results in overfitting and exponentially growing computational cost.  
It also renders the networks to being more data-driven which makes them unable to work under a small change in the point set.
Although several works~\cite{zhi2017lightnet,ma2017bv,liu2019relation} show less tendency towards overfitting, their performance is highly dependent on the samples.
Most models are susceptible to irregular samples and work only under certain conditions. 
Meanwhile, developing highly accurate, robust and fast models for the processing of 3D data is demanded by many practical applications like autonomous driving.

Our studies have shown that patterns of an object represent the structural information of that object and remain almost unchanged under different samplings of the data and so are a more stable basis on which to build classification and segmentation algorithms.
In this paper, we propose a novel cloning technique aiming at extraction of the stable patterns from an object. 
Such features can efficiently improve the network's performance. Our robust network is the first successful attempt to tackle and to investigate classification and segmentation under irregular samplings of point cloud data. Additionally, it needs a relatively low number of parameters so it is fast and less prone to overfitting. The key contributions of this paper are as follows:
\begin{itemize}
   \item We design a robust deep neural network whose performance is not significantly affected by data grid, thus it is invulnerable to noise, out-of-distribution samples and distortions.
   \item The proposed model mitigates the problem of distance saturation in KNN-based models.
   \item The architecture of the proposed network allows the user to go deeper and deeper without the problem of vanishing-gradient. This scheme is capable of analysis of highly complex objects. 
   \item We provide thorough empirical and theoretical analysis on the stability and efficiency of the proposed method using the 3D benchmark datasets: ModelNet and ShapeNet \cite{wu20153d,yi2016scalable}.
\end{itemize}
In order to be robust to undesirable factors, our pattern-based network does not require any annotated samples, and it just trains once on noise-free point cloud data and then runs over any distorted ones\footnote{Supplementary materials are available at \url{https://github.com/mogvision/pattern-net}}.% To facilitate future research, we make the code publicly available at \url{https://github.com/mogvision/pattern-net}.???
%##########################################
\section{Related work}
Deep learning methods for 3D shape analysis and understanding can be broadly divided into view-based, volumetric and point cloud-based categories.
View-based techniques~\cite{su2015multi,qi2016volumetric,wang2019dominant} map 3D models into 2D view scenes and then employ image-based CNNs for further analysis. Self-occlusions and information loss often occur during mapping. Volumetric methods~\cite{maturana2015voxnet,wu20153d,riegler2017octnet,tatarchenko2017octree,klokov2017escape} quantize the input 3D models into a regular grid before applying 3D CNNs. Loss of resolution, high memory and computational demands are the main limitations of voxelization.

Instead of converting or mapping 3D models into other domains, they can be analyzed in the point cloud domain directly.
Due to the impressive results of PointNet~\cite{qi2017pointnet}, most studies have been devoted to learning directly in the point cloud domain.
Since PointNet does not consider the local pattern of a given 3D point cloud, PointNet++~\cite{qi2017pointnet++} was proposed, which uses a hierarchical application of PointNet to multiple subsets of a 3D point cloud. 
Inspired by DenseNet \cite{huang2017densely}, DensePoint \cite{liu2019densepoint} was introduced that learns a dense contextual representation for point cloud processing via a deep hierarchy architecture.
Exploitation of other aspects of local structure  with PointNet are
also reported in~\cite{guerrero2018pcpnet,shen2018mining}.
Superpoint~\cite{landrieu2018large} partitions the point cloud into geometrical homogeneous elements and then a graph convolution network is applied to such local elements.
The main drawback of these methods is their lack of shape awareness. 
More precisely, such methods do not explicitly model the local spatial layout of points. To this end, several works have been developed~\cite{liu2019relation,wang2019dynamic,xu2019grid,zhang2019rotation} that capture the spatial layout of a point cloud by learning a high-level relation expression among 3D points.

Although the approach of explicitly modeling the relation improves the segmentation results, isolating high-level relation features from low-level ones may not strengthen relation propagation, subsequently there is a vanishing-gradient problem~\cite{huang2017densely}.
The appropriate distance or neighbour count parameter to use in the KNN-based networks~\cite{liu2019relation,wang2019dynamic,liu2019densepoint} is often obtained via trial-and-error, which is saturated at a certain number of neighbours and even drastically drops by increasing the number of neighbours or neighbourhood radius. This aspect of such networks is not favorable, especially in dense 3D models.

Another main weakness of the existing networks that makes them impractical for real world data is their vulnerability to noise, distortions and out-of-distribution sampling schemes.
Robustness is a key property that allows applying the same model to different irregular point clouds. Enriching the training step with annotated data, like adding noise, distortions etc, is not a straightforward solution to the above problem. This is because increasing the volume of input data with annotated samples makes the training convergence difficult to achieve. Additionally, a large-scale input data renders a high number of parameters, resulting in overfitting.
Hence there is interest in developing CNNs that work efficiently under disparity between the training and test data.
%--------------------------------------------
\section{Proposed Pattern-Wise Network}\label{sec::proposed}
The goal is to establish and learn a deep neural network that converts an input point cloud $P=\{p_m\in\mathbb{R}^d, m=1,...,M\}$ into a set of segmentation labels $\Gamma_s=\{\gamma_s\in\mathbb{R}, s=1,...,S\}$ or a set of classification labels $\Gamma_c=\{\gamma_c\in\mathbb{R}, c=1,...,C\}$. 
Here, $M$ is the total number of 3D points and $d$ is the dimension of the point set that can be represented as a set of 3D coordinates plus other measured features like color, normal vectors etc $p_m=\{coordinate: (x_m,y_m,z_m), color:(r_m,g_m,b_m), normal: (N^x_m,N^y_m,N^z_m)\}$.
In this paper, we just consider the 3D coordinates and extension of the network over color and normal is straightforward.
The output of the network for segmentation tasks is a vector of labels $\gamma_s\in\{1,...,S\}$, where $S$ is the number of segmentation labels. Likewise, for the classification task, 3D points are labelled as $\gamma_c\in\{1,...,C\}$ with $C$ classes.
%--------------------------------------------
\subsection{Network Properties}\label{sec::propweties}
A segmentation/classification network for a point set must meet the following four requirements~\cite{qi2017pointnet,qi2017pointnet++,atzmon2018point}:

\textbf{Property i} (\textit{Permutation Invariant}): It states that segmentation/classification scores must be invariant to changes in the order of 3D points. Unlike pixels in images or voxels in volumetric grids, 3D point cloud has no order and due to its irregular format, the network must be invariant to the order of points.

\textbf{Property ii} (\textit{Transformation Invariant}): The labels/classifications of points must not be varied by their changes in rotation, scale and translation.

\textbf{Property iii} (\textit{Points Relations}): In the point cloud, the relation between points is determined by their distance from each other.
The distance metrics could be Euclidean distance, Manhattan distance, cosine distance, etc. Points in the point cloud are not isolated and their neighbours make a meaningful subset that can be measured by an appropriate metric.

\textbf{Property iv} (\textit{Robustness}): 
The segmentation parts or classification labels of points must not be varied under different samplings. In practice, the distribution of test data is not close to that of the training data. Additionally, presence of out-of-distribution samples, noise and distortions in test data is inevitable so point cloud networks must be robust to the irregular samples.

The above properties are the backbone of our network.
\subsection{Network Architecture}\label{sec::arch}
\begin{figure*}[!ht]%##############################################
\centering
    \subfigure[Cloning analysis]{ \includegraphics[width=4.2in]{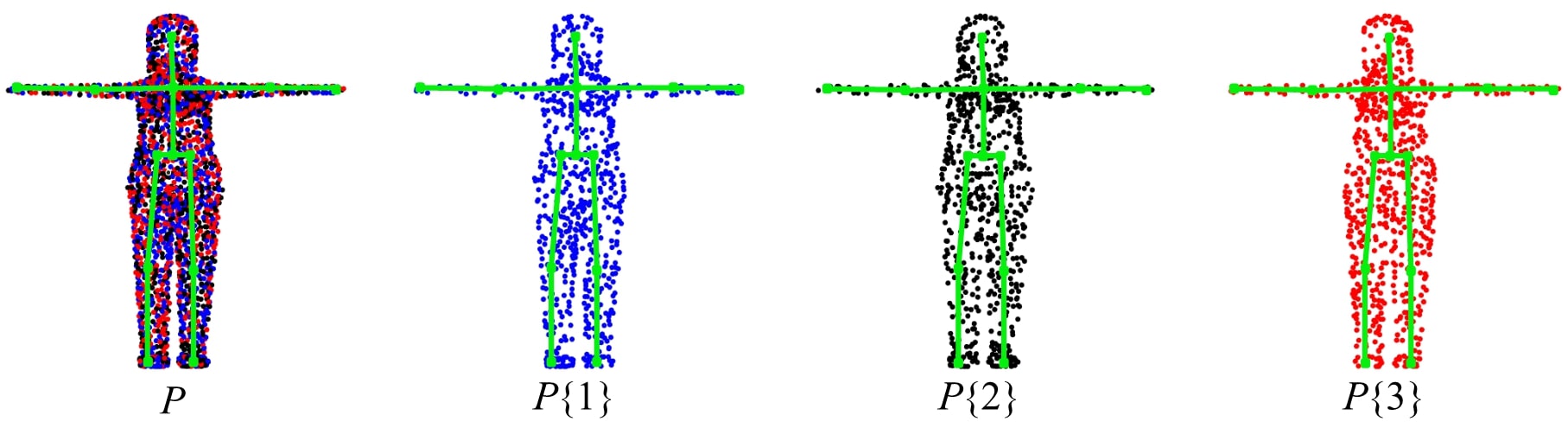}\label{fig::msa}}
    \subfigure[Entropy of patterns ]{\includegraphics[width=1.55in]{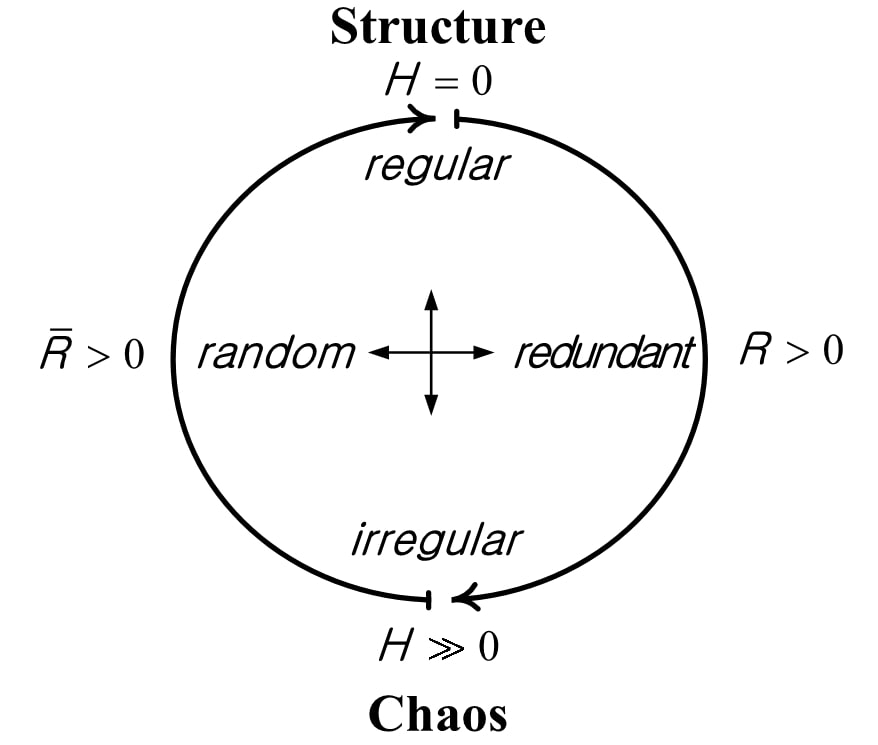}\label{fig::msb}}
    \subfigure[Categories of patterns ]{\includegraphics[width=1.9in]{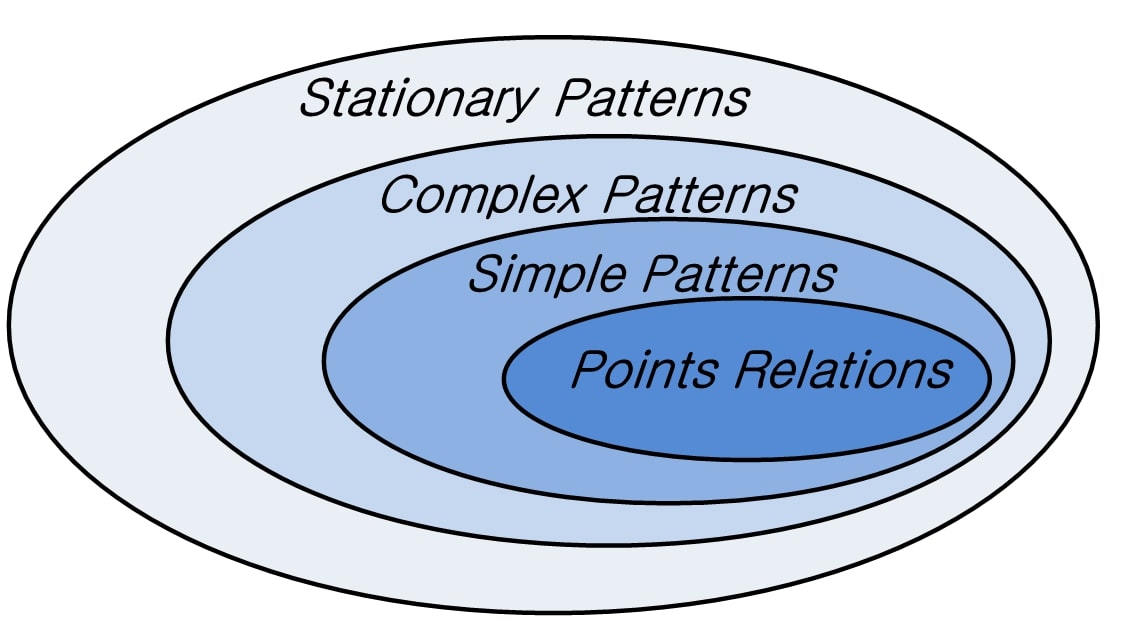}\label{fig::msc}}
\caption{(a) $P$ is the input point set which is constituted of three subsets shown in blue ($P\{1\}$), black ($P\{2\}$) and red ($P\{3\}$). The abstract/structural information (solid green lines) remains unchanged across different samplings. (b) Entropy of patterns with different characteristics. (c) Categorization of potential patterns inside a rigid/non-rigid object in point clouds.}
\end{figure*}%########################################

The orderly disorder theory was introduced in physics~\cite{peters1996chaos,davood2015orderly} and it refers to a way of describing the complex structure of disorders within complex systems.
Unpredictable disorders could occur just under external disturbances not because of internal reasons.
Ordered/predictable disorders may not be seen by human vision and this increases the ambiguity between the predictable and unpredictable disorders. However, the entropy metric could give us the degree of chaos inside a complex structure.
Chaos theory has been well studied in mathematics, behavioral science, management, sociology etc. With the success of CNNs in solving high-order problems, our aim is to deeply analyze the links between points in the given point cloud.

For a rigid/non-rigid 3D object in the given point cloud, the location of points may change under different sampling operations, external disturbances etc. but its abstract information does not vary. Such abstract information can be named as \textit{stationary} information and they are predictable while \textit{non-stationary} information refers to those features that do not obey a regular pattern and under different conditions show different behaviours. The best way is to encode a 3D object by its stationary patterns which always show predictable behaviour. Contrary to stationary patterns, the behaviour of non-stationary patterns is completely unpredictable. For example, the minor details of a 3D model can vary under various samplings but its skeleton remains unchanged [Fig. \ref{fig::msa}]. Thus, we can claim that the classification score of a 3D object must not be varied under changes in the density and distribution of points if the number of points is sufficiently large, i.e.
\begin{equation} \label{eq::scale}
    \Gamma_{[p_1,...,p_N]} = \Gamma_{[p_1,...,p_M]}\qquad if \qquad N < M\ \&\ N\gg1.
\end{equation}
In other words, if our observation of an object is sufficiently detailed, then few sample variations must not change the label of the enquiry object.
Let's assume $M=LN$ for a sufficiently large $N$, where $L$ is a positive integer. According to Eq. \ref{eq::scale}, we can say that
\begin{equation} \label{eq::scale2}
\Gamma_{[p_1,...,p_N]} = \Gamma_{[p_{(l-1)N+1},...,p_{lN}]} 
    ,\quad \forall l \in \{2,...,L\}.
\end{equation}

One possible solution to the above equation is to decompose the input point cloud $P$ into $L$ levels via a random down-sampling operator in such a way that all $L$ point subsets $P{\{l\}},l\in\{1,...,L\}$, are completely different while their overall schemes/abstracts are similar to each other. Under these conditions, Eq. \ref{eq::scale2} is asserted. 
%Let function $f$ denote the structural information of samples in the point cloud domain.
If we apply a random down-sampling operator to point cloud $P$ that provides
\begin{align} \label{eq::scale3}
    &P{\{l\}} \cap P\{j\} = \emptyset \quad \forall \,l,j \in \{1,...,L\}\ \&\  l\neq j,
\end{align}
\begin{align} \label{eq::scale4}
    &\bigcup^L_{l=1}P{\{l\}} = P,
\end{align}
\begin{align} \label{eq::scale5}
    &H(P{\{l\}}) \simeq H(P{\{j\}}) \quad   \quad \forall\, l,j \in \{1,...,L\}\ \&\  l\neq j,
\end{align}
then we can assert that all the $L$ point subsets have similar \textit{stationary} structures/patterns. In Eq. \ref{eq::scale5}, `$H$' denotes the entropy of each subset and this equation assures that all the subsets have approximately similar information content. 
If $R$ denotes the entropy of redundant/predictable patterns in a point cloud set and similarly, the entropy of random/less-redundant/unpredictable ones is denoted by $\overline R$, then the entropy $H$ of each subset is equal to 
\begin{align} \label{eq::scale6}
     H\big(P{\{l\}}\big) = R\big(P{\{l\}}\big)+ \overline R\big(P{\{l\}}\big), \quad \quad \forall\, l \in \{1,...,L\}.
\end{align}
\begin{figure*}[!ht]%########################################
\centering
    \subfigure[$P$]{
    \includegraphics[width=1.4in]{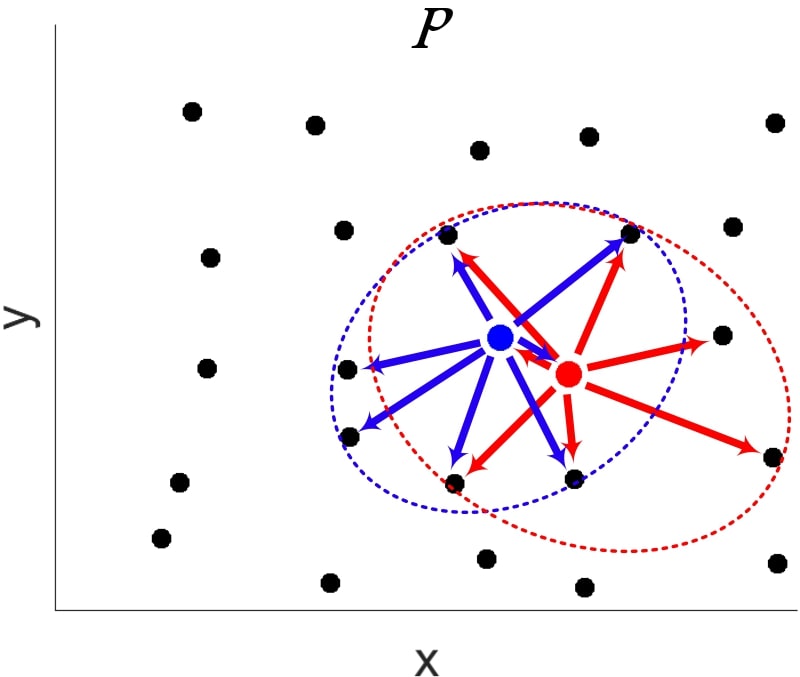}
    \label{fig::asym_symA}}
    \subfigure[$P{\{1\}} \cap P\{2\} = \emptyset$ and $P{\{1\}} \cup P\{2\} = P$]{
    \includegraphics[width=1.42in]{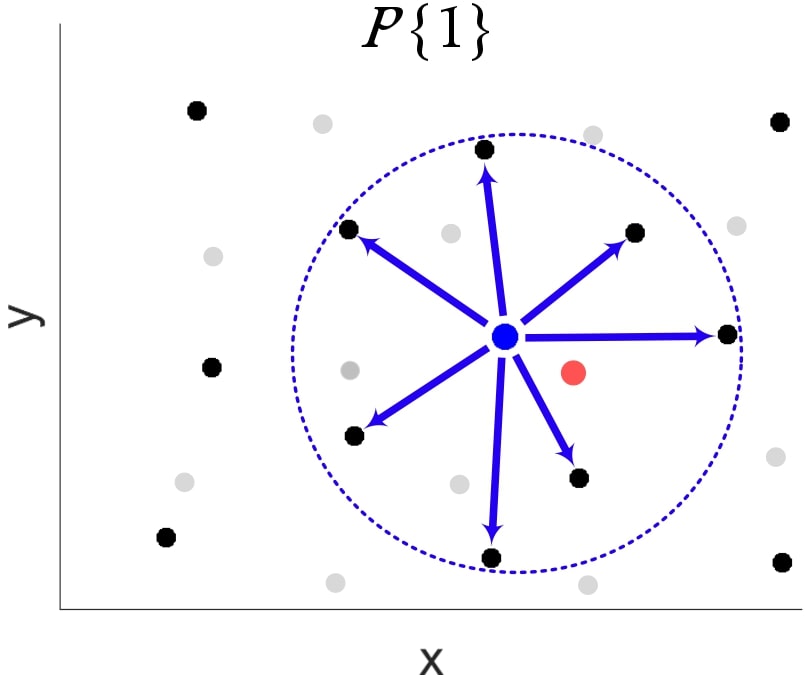}
    \includegraphics[width=1.42in]{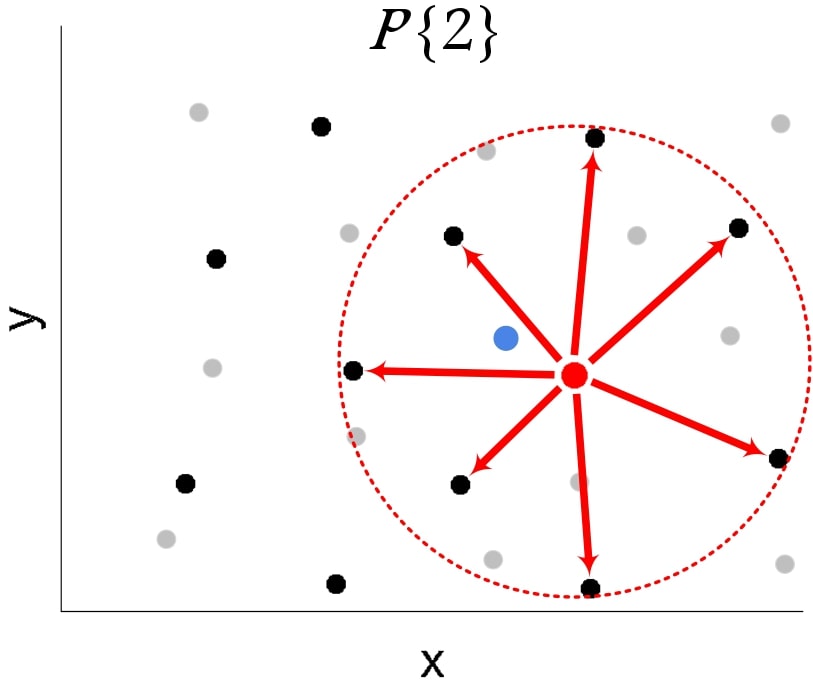}
    \label{fig::asym_symB}}
\caption{An example of how the proposed cloning decomposition can successfully remedy chaos in the given point cloud. For better visualisation, axis `z' was set to 0.
(a) Conventional KNN-based networks yield asymmetric, complex, and dissimilar patterns for two adjacent points; (b) The cloning decomposition technique increases the probability of extracting symmetric, simple, and similar patterns.}
\label{fig::asym_sym}
\end{figure*}%########################################
The entropy of redundant patterns is close to zero and the large part of the overall entropy `$H$' is assigned to unpredictable patterns $\overline R$. Possible values of entropy for patterns with different characteristics is depicted in Fig.~\ref{fig::msb}.
It is worth noting that the \textit{randomness} of the down-sampling operator ensures that each point subset includes all parts/organs of an object. 
We name this strategy as `cloning decomposition' and an example is illustrated in Fig. \ref{fig::msa}, where all the point subsets of the given 3D object share approximately similar structures across multiple decomposed levels while none of them shares identical 3D samples.
Overall, the feature space of an object could be categorized into a number of undesirable non-stationary patterns in which the patterns are chaotic and a series of stationary patterns, wherein complex and simple patterns are learnt from orderly stable relations [Fig. \ref{fig::msc}].

The simplified point subsets can markedly help the network to extract stationary patterns, enhancing robustness as stated by \textbf{Property iv} above. 
An illustrative example is depicted in Fig. \ref{fig::asym_sym}. This example aims at drawing the KNN responses (here, $K=7$) of two adjacent points. According to Fig. \ref{fig::asym_symA}, conventional KNN-based networks rely heavily on the density of 3D points and also they may not find reliable neighbours as the radius of the neighbourhood for an enquiry point depends on the density and distribution of the point cloud data. 
The proposed technique decomposes the given point cloud in Fig. \ref{fig::asym_symA} into two subsets via Eqs. \ref{eq::scale3}-\ref{eq::scale5}.
As shown in Fig. \ref{fig::asym_symB}, it could provide similar patterns for a complex point cloud. The radius of the neighbourhood for both the blue and the red points is almost equal.
As will be discussed later, this strategy efficiently helps the network not to be saturated with its K nearest neighbours while keeping the radius of the neighbourhood reasonable.

\begin{figure*}[!ht]%###################################
\centering
	\includegraphics[width=\linewidth]{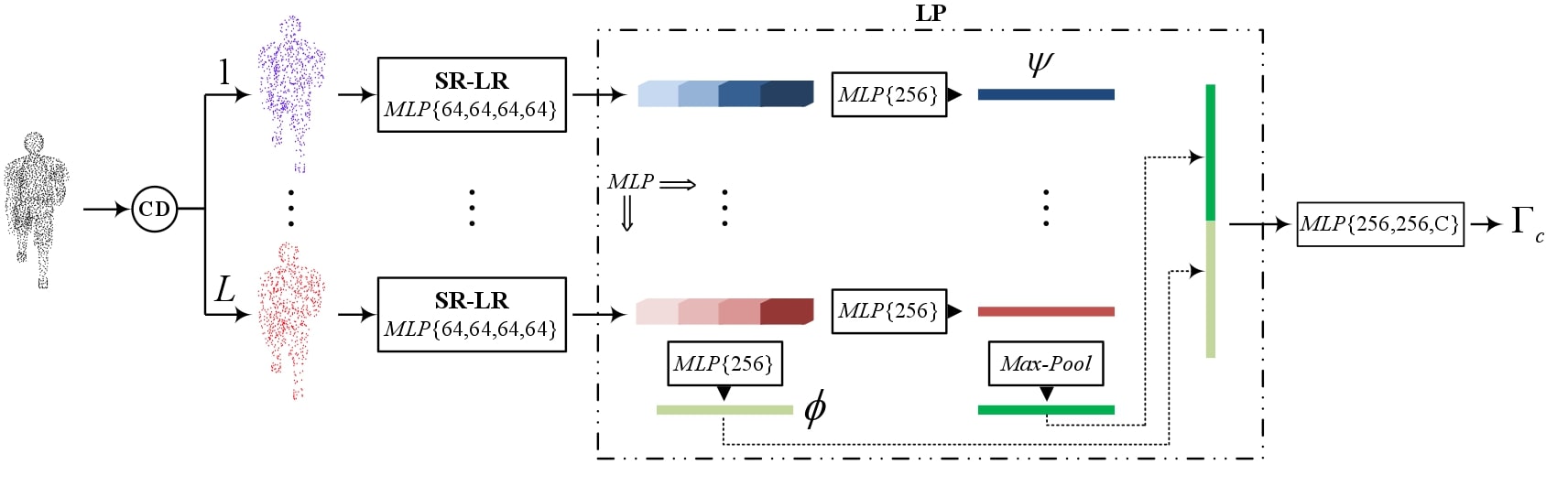}
\caption{Proposed network architecture for classification of point cloud data.}
\label{fig::network}
\end{figure*}

The proposed network is depicted in Fig.~\ref{fig::network}. The framework has four main layers including cloning decomposition (CD), searching relations (SR), learning relations (LR) and linkage patterns (LP) layers. In the following, we detail the functionality of each layer.
\begin{itemize}
    \item \textbf{Cloning Decomposition (CD) Layer}: Image acquisition is often made under various conditions that directly affect quality and quantity of 3D models and subsequently, their point cloud samples.
    This layer decomposes the input object points into multiple subsets via Eqs. \ref{eq::scale3}-\ref{eq::scale5}. Since high-level patterns are directly deduced from the low-level ones, this layer plays a key role in the extraction of reliable patterns.
    Cloning analysis reduces the number of parameters, alleviates the vanishing-gradient problem and improves the convergence pace.
\end{itemize}
%########################################
\begin{itemize}   
    \item \textbf{Searching Relations (SR) Layer}: The task of this layer is to search all possible links between the feature vectors (or equivalently potential \textit{low-level} patterns) according to \textbf{Property iii} above. This is done via the KNN algorithm in the Hilbert space. Instead of using the Euclidean space dominantly employed by most existing deep networks, we use a Hilbert kernel. Given two low-/high-level feature vectors $X$ and $Y$, a measure/relation of similarity between them in the Euclidean space can be expressed as 
    \begin{align} \label{eq::dist1}
    \mathcal{E}(X,Y)=||X||^2_2+||Y||^2_2-2 <X,Y>,
    \end{align}
    while the Hilbert kernel yields 
    \begin{align} \label{eq::dist2}
      \mathcal{H}(X,Y)=2\big(1-<\frac{X}{||X||_2},\frac{Y}{||Y||_2}>\big).
    \end{align} 
    In the above equations, the angle brackets denote an inner product operator. The Hilbert kernel emphasizes the cross-similarity while the self-similarity remains the same, i.e. unit. Unlike the Euclidean space that is biased by self-similarity, the Hilbert space just considers the cross-similarity between enquiry feature vectors and thus, it is expected that it can find reliable relations between feature vectors.
\end{itemize} 
\begin{itemize}    
    \item \textbf{Learning Relations (LR) Layer}: This layer seeks and learns potential relations between all input feature vectors via a convolution kernel followed by a batch normalization operator. A max pooling operator is then applied to the outputs to obtain the global feature of the input features. The max pooling is a symmetric function that guarantees that the extracted features are permutation-invariant, as stated by \textbf{Property i} above.
    The combination of the convolution kernel, batch normalization and max pooling operators is often called multi-layer perceptron (MLP) \cite{qi2017pointnet}.
\end{itemize}
\begin{itemize}   
    \item \textbf{Linkage Patterns (LP) Layer}: 
    This layer comprises several MLP operators and its aim is to aggregate the relations across all the subsets and extract the most stable patterns from them.
    All the subsets have different samples and the LP layer is applied to the patterns for extracting the common ones. Such patterns carry stationary information of the object and they are robust to irregular samples and density.
\end{itemize}
Now, we detail the proposed pattern-based CNN (hereafter it is abbreviated as Pattern-Net) for the classification and segmentation of given point clouds.
%###############################################################
\subsection{Classification and Segmentation Networks}\label{sec::cls_seg}
In the classifier model depicted in Fig.~\ref{fig::network}, the input point cloud data is first decomposed into `$L$' subsets via the cloning technique described in the previous section. Inside each subset, relations between each query point and its neighbours is sought by the KNN algorithm. This is done by applying four MLPs $\{64,64,64,64\}$ to each cloning subset.
Similar to KNN-based networks~\cite{li2018so,wang2019dynamic,liu2019relation}, we compute K nearest neighbour responses of edges emanating from the enquiry feature point and stack them with the enquiry feature point.
The KNN algorithm is considerably affected by the density of points. If the 3D model is sparse, then the best K responses will lie within a large volume neighbourhood while such responses in a high density model may fall into a small radius [Fig.~\ref{fig::asym_symA}]. 
Moreover, adjacent points in high density 3D samples would share similar KNN responses and this makes the network data-dependent. In other words, the network may fail to work for test point cloud data with different samplings from the training data. 
This problem can be seen in almost all KNN-based networks which here is solved by the cloning layer.
In the proposed model, the LP layer finds the most repeated relations across different subsets and then labels them as stable patterns. 
Even in the presence of changes in point coordinates, the patterns remain approximately unchanged as we consider overall behaviour of a group of points rather than the exact behaviour of each point.

Each cloning subset yields a description vector of length 256 and they are called cloning description vectors $\psi_l, l \in\{1,...,L\}$. All the $L$ cloning description vectors are arranged in a matrix $\Psi$.
An MLP is applied to all cube features over all the subsets to yield a global description vector $\phi$, shown in light green in the figure. Here, the goal is to make each cloning description vector $\psi_l$ similar to the global one $\phi$ as much as possible. If we consider a linear relationship between the cloning and global description vectors, i.e. $\phi=\Psi \omega$, then the estimated coefficients $\omega$ can be computed by the Moore–Penrose inverse, i.e. $\omega =\Psi^{\dagger} \phi= (\Psi^T\Psi)^{-1}\Psi^T\phi$. Parameter $\omega$ determines the contribution of each cloning vector in the resultant global vector.
The deviation of $\omega$ elements should approach zero if all the cloning description vectors are completely similar to the global one. We add this term into the loss function:
\begin{equation}\label{eq::loss}
    L(\theta)=\underbrace{-\frac{1}{n} \sum_{i=1}^{n}\sum_{c=1}^{C} y_{ic}\log{p_{ic}}}_{classification\:loss}+ \lambda\underbrace{\sigma(\Psi^{\dagger}\phi)}_{linear\: mapping\:loss}
  \end{equation}
%##############################################
\setlength{\tabcolsep}{6.0pt} % Default value: 6pt
\begin{table}
\caption{Classification accuracy in percentage (\%) on ModelNet40 (`-':unknown)} \label{tab:cls} 
\centering 
\begin{tabular}{l ccc} 
\toprule % Top horizontal line
Method
&Input
& Avg. classes & Overall\\
\midrule % In-table horizontal line
PointNet \cite{qi2017pointnet}&  1k-xyz& 86.0 & 89.2 \\
PointNet++ \cite{qi2017pointnet++}& 5k-xyz&  - & 91.9 \\
PointCNN\cite{atzmon2018point} & 1k-xyz& 88.1 & 92.2\\ 
ECC\cite{simonovsky2017dynamic} & 1k-xyz& 83.2 & 87.4\\ 
DGCNN\cite{wang2019dynamic} & 2k-xyz& 90.7 & 93.5 \\ 
SO-Net\cite{li2018so} & 2k-xyz& 88.7 & 90.9  \\ 
DensePoint\cite{liu2019densepoint}&1k-xyz& - & 93.2\\
RS-CNN\cite{liu2019relation} & 1k-xyz& - & 93.6\\ 
\midrule 
        & 1k-xyz&90.3 & 92.9 \\ 
Pattern-Net& 2k-xyz&90.7 & 93.6 \\ 
        & 4k-xyz&\bf{90.8} & \bf{93.9} \\ 
\bottomrule % Bottom horizontal line
\end{tabular}
\end{table}%##############################################
In the above equation, the first term is the categorical cross-entropy function for computing the loss of the predicted labels and the second term enforces the network to yield zero standard deviation for weights obtained by the linear mapping between the cloning and global description vectors. $\lambda$ is a predetermined constant, whose value is determined by the smoothing label's value in the one-hot encoded $y_{ic}$. Finally, $p_{ic}$ is a scaled (softmax) logits. The Moore–Penrose pseudo-inverse can be simply implemented by singular value decomposition (SVD) \cite{brake2019singular}.
The ultimate cloning vector is obtained by applying a max-pooling operator to the cloning vectors. The resulting vector is aggregated with the global vector to yield the description vector for the given point set which is of length 512. 
Finally, the description vector is fed into three MLPs $\{256,256,C\}$, configured for classification.
The resultant description vector is also used in segmentation. For this task, four MLPs \{64,64,64,64\} are applied to the input data to extract low- and high-level features. They are then concatenated with the description vector to encode each point. Similar to the classification task, three MLPs $\{256,256,S\}$ are employed for segmentation. The drop-rate of all decoding MLPs except the last one is fixed at $0.5$. 
%-------------------------------------------------
\section{Experimental Results}
%########################################
\begin{figure}[!ht]
\centering
    \subfigure[]{\includegraphics[width=0.43\linewidth]{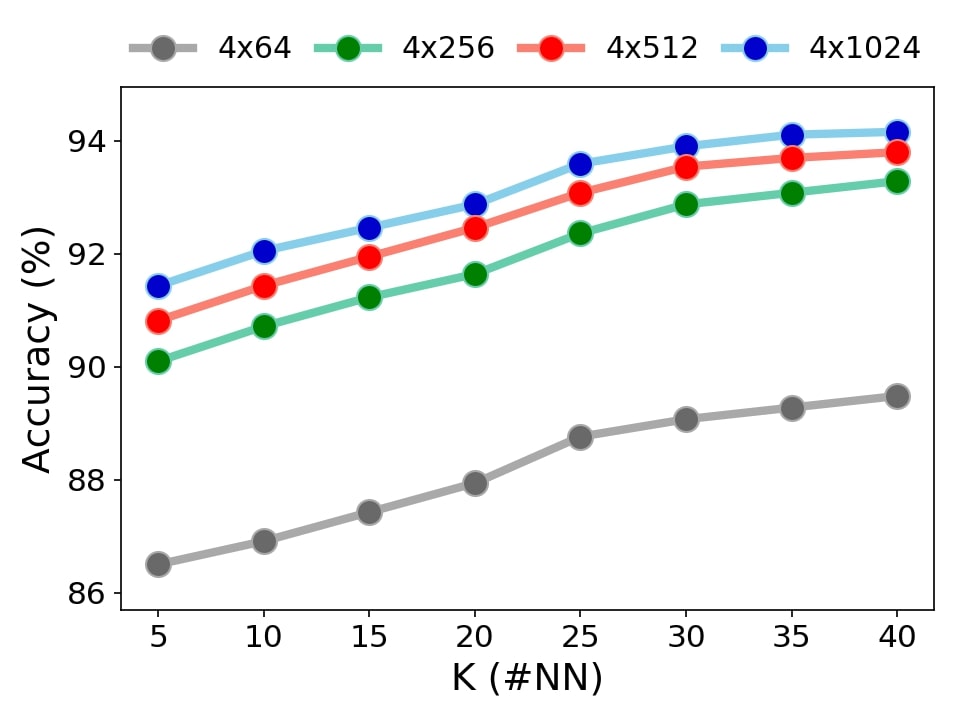}\label{fig::knn}}
    \subfigure[]{\includegraphics[width=0.53\linewidth]{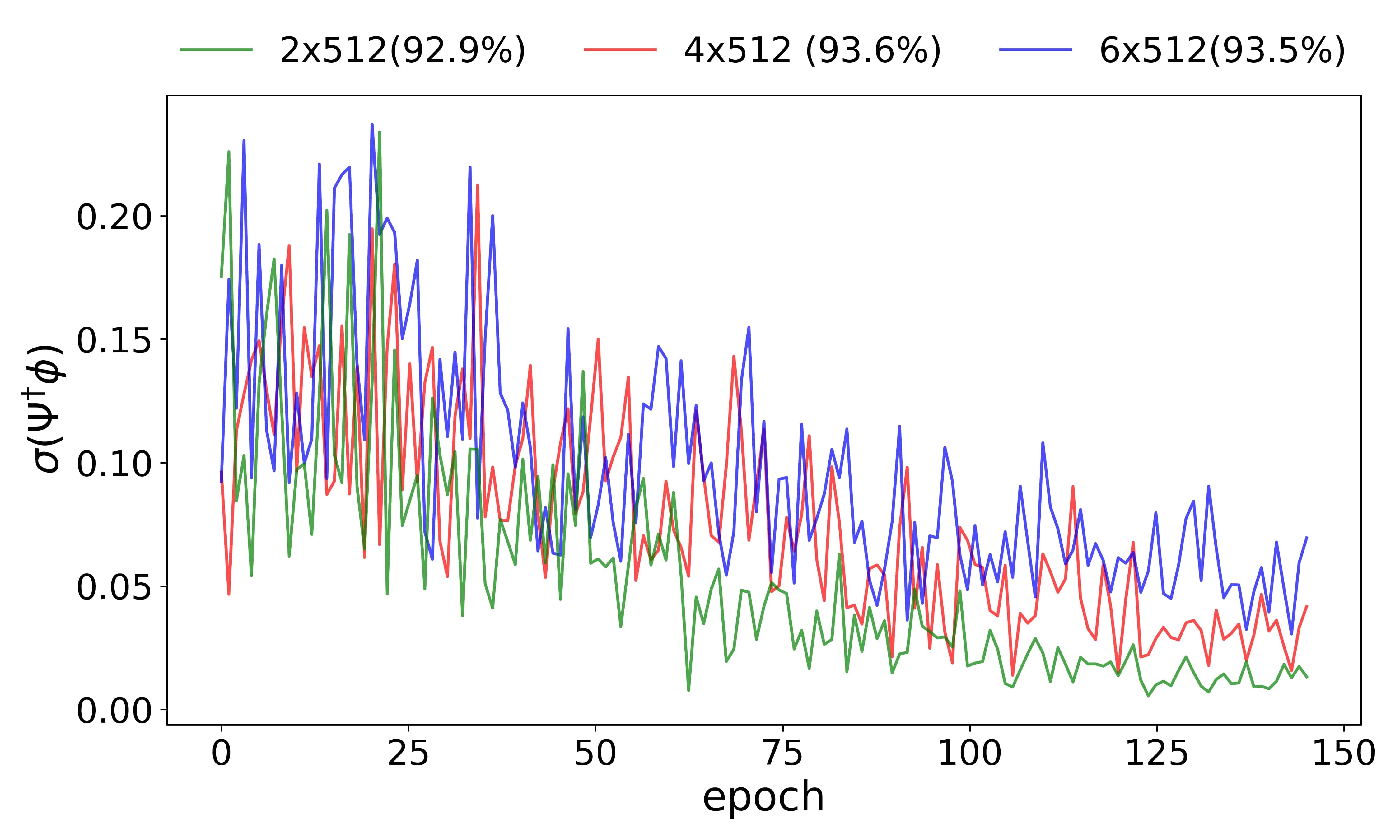}\label{fig::error}}
\caption{(a) Influence of parameter K and number of points on the classification accuracy. (b) Linear mapping loss for different cloning levels (\#cloning levels$\times$\#points).}
\end{figure}%########################################
\setlength{\tabcolsep}{8pt} % Default value: 6pt
We have evaluated Pattern-Net on the ModelNet40 dataset \cite{wu20153d} for the classification task. It contains 12311 meshed CAD models from 40 categories. Similar to the other work, 9843 models were used for training and the rest for testing and the models were normalized to a unit sphere. Each model is uniformly sampled from the mesh faces in 1k, 2k and 4k samples. During the training step, the points are augmented by randomly rotating, scaling and translating for being transformation invariant (\textbf{Property ii} above). The quantitative comparisons with the state-of-the-art point-based methods are presented in Table \ref{tab:cls}. Our method for 1k and 2k points is on par with the other methods and gives the best result for 4k points. 
\setlength{\tabcolsep}{5pt}%########################################
\begin{table*} 
\caption{Classification accuracy in percentage (\%) on ModelNet40 in the presence of noise [$2k-xyz+\mathcal{N}(0,\sigma)$]} %\label{table:complexity} 
\centering 
\begin{tabular}{l cc cc cc cc cc} 
\toprule % Top horizontal line
{Method} 
&\multirow{2}{*}{}
& {$\mathcal{N}(0,0.02)$}
& {$\mathcal{N}(0,0.05)$}
& {$\mathcal{N}(0,0.08)$}
& {$\mathcal{N}(0,0.1)$} 
& {$\mathcal{N}(0,0.15)$}\\
\midrule % In-table horizontal line
PointCNN &&  78.7 &  40.8 & 18.6 & 10.5 & 4.7 \\ 
DGCNN  &&  92.9 &  69.1 & 29.9 & 11.4 & 4.2 \\ 
SO-Net && 70.6 & 35.4 & 11.9 & 9.8& 5.8\\
\midrule
Pattern-Net(4x512) &&  \bf{93.5} & \bf{92.4} & \bf{89.1} & \bf{84.2} & \bf{32.6}  \\
\bottomrule % Bottom horizontal line
\label{table::noise}
\end{tabular}
\end{table*}%############################################
Unlike the existing methods, where the performance is quickly saturated by a specific number of points, our network is saturated slowly as can be seen in Fig. \ref{fig::knn}. For a fixed number of decomposed levels, an increase in the number of points improves the classification accuracy. There is also a direct link between the accuracy and parameter K in KNN. Tuning parameter K in the KNN-based networks is not straightforward and it is often obtained by trial-and-error. According to the figure, further increasing K yields better results but the  performance increase slows down after $K=30$. In order to make a balance between computational time and performance, parameter K was set to 30 throughout this study. 
\begin{figure}[!ht]%#######################################
\centering
\includegraphics[width=0.95\linewidth]{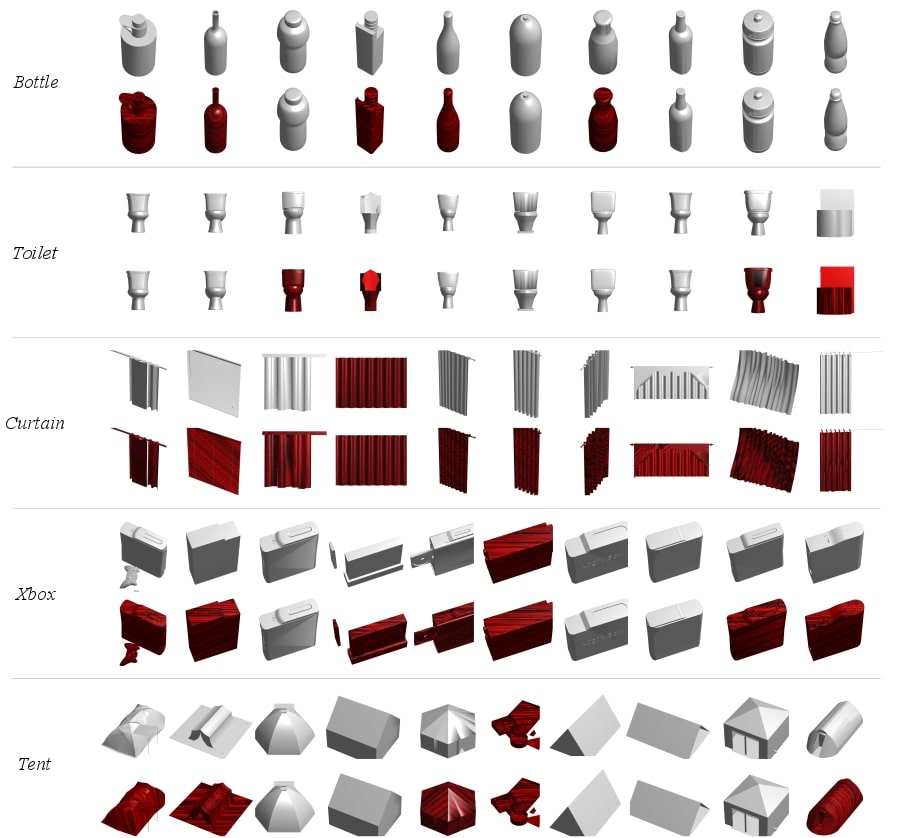}\label{fig::cls_results}
\caption{Classification results on ModelNet40 with added Gaussian noise $\mathcal{N}(0,0.05)$. First 10 shapes 
shown are for each query, with the first line for our Pattern-Net and the second line for DGCNN. The misclassified objects are highlighted in red.}
\end{figure}%#######################################

A suitable value for the number of cloning levels relies on the number of input points, parameter K and the convergence of Eq. \ref{eq::loss}. As shown in Fig. \ref{fig::error}, the rate of convergence is reduced by increasing cloning numbers  and the network requires more epochs to find similar patterns between different cloning levels. However, lower mapping loss does not guarantee high accuracy as the loss equation is constituted of two terms including the mapping loss and the classification loss. Our experiment shows that \{3,4,5\} cloning levels often yield good results. In Table \ref{tab:cls}, this parameter was set to 4.

As mentioned previously, the ability to tolerate noise is a necessity for robust and practical deep learning methods. In the following experiment, we added zero-mean white Gaussian noise with different standard deviation values $\mathcal{N}(0,\sigma)$ to the test samples. The classification accuracy results are reported in Table \ref{table::noise} and a part of results is depicted in Fig. \ref{fig::cls_results}. The table shows that our networks can tolerate noise up to $\mathcal{N}(0,0.1)$ while the state-of-the-art methods failed to work. The drop is less than 10\% for $\mathcal{N}(0,0.1)$, which is impressive. \\

\setlength{\tabcolsep}{1.0pt}%############################################
\begin{table*} 
\caption{Segmentation results (\%) on ShapeNet}
\centering 
\begin{tabular}{l|c ccccccc } 
\toprule % Top horizontal line
Category (\#)&
    & PointNet & PointNet++
    & PointCNN & DGCNN
    & SO-Net & RS-CNN & Pattern-Net\\       
\midrule
Areo (2690)&&   83.4 & 82.4 & 82.4 & 84.0 & 82.8 & 83.5 & \bf{84.3}\\ %
Bag (76)&&      78.7 & 79.0 & 80.1 & 83.4 & 77.8 & \bf{84.8} & 81.0\\ %
Cap (55)&&      82.5 & 87.7 & 85.5 & 86.7 & 88.0 & \bf{88.8} & 87.4\\ %
Car (898)&&     74.9 & 77.3 & 79.5 & 77.8 & 77.3 & 79.6 & \bf{80.1}\\
Chair (3758)&&  89.6 & 90.8 & 90.8 & 90.6 & 90.6 & 91.2 & \bf{91.4}\\
Ear (69)&&      73.0 & 71.8 & 73.2 & 74.7 & 73.5 & \bf{81.1} & 79.7\\
Guitar (787)&&  91.5 & 91.0 & 91.3 & 91.2 & 90.7 & \bf{91.6} & 91.4\\
Knife (392)&&   85.9 & 85.9 & 86.0 & 87.5 & 83.9 & \bf{88.4} & 88.1 \\
Lamp (1547)&&   80.8 & 83.7 & 85.0 & 82.8 & 82.8 & 86.0 & \bf{86.3}\\
Laptop (451)&&  95.3 & 95.3 & 95.7 & 95.7 & 94.8 & \bf{96.0} & 95.8\\
Motor (202)&&   65.2 & 71.6 & 73.2 & 66.3 & 69.1 & \bf{73.7} & 72.1\\
Mug (184)&&     93.0 & 94.1 & 94.8 & \bf{94.9} & 94.2 & 94.1 & 94.1\\
Pistol (283)&&  81.2 & 81.3 & 83.3 & 81.1 & 80.9 & \bf{83.4} & 82.2\\
Rocket (66)&&   57.9 & 58.7 & 51.0 & \bf{63.5} & 53.1 & 60.5 & 62.4\\
Skate (152)&&   72.8 & 76.4 & 75.0 & 74.5 & 72.9 & \bf{77.7} & 72.4\\
Table (5271)&&  80.6 & 82.6 & 81.8 & 82.6 & 83.0 & 83.6 & \bf{83.9}\\
\midrule % In-table horizontal line
Avg. && 83.7 & 85.1 & 85.1 & 85.2 & 84.9 &86.2 & \bf{86.4}\\ %   1st% x 1st# + 2nd% x 2nd# / 1st#+2st#
\bottomrule % Bottom horizontal line
\end{tabular}
\label{table::seg_res}
\end{table*}%############################################
\begin{figure}[!ht]%############################################
\centering
\includegraphics[width=0.95\linewidth]{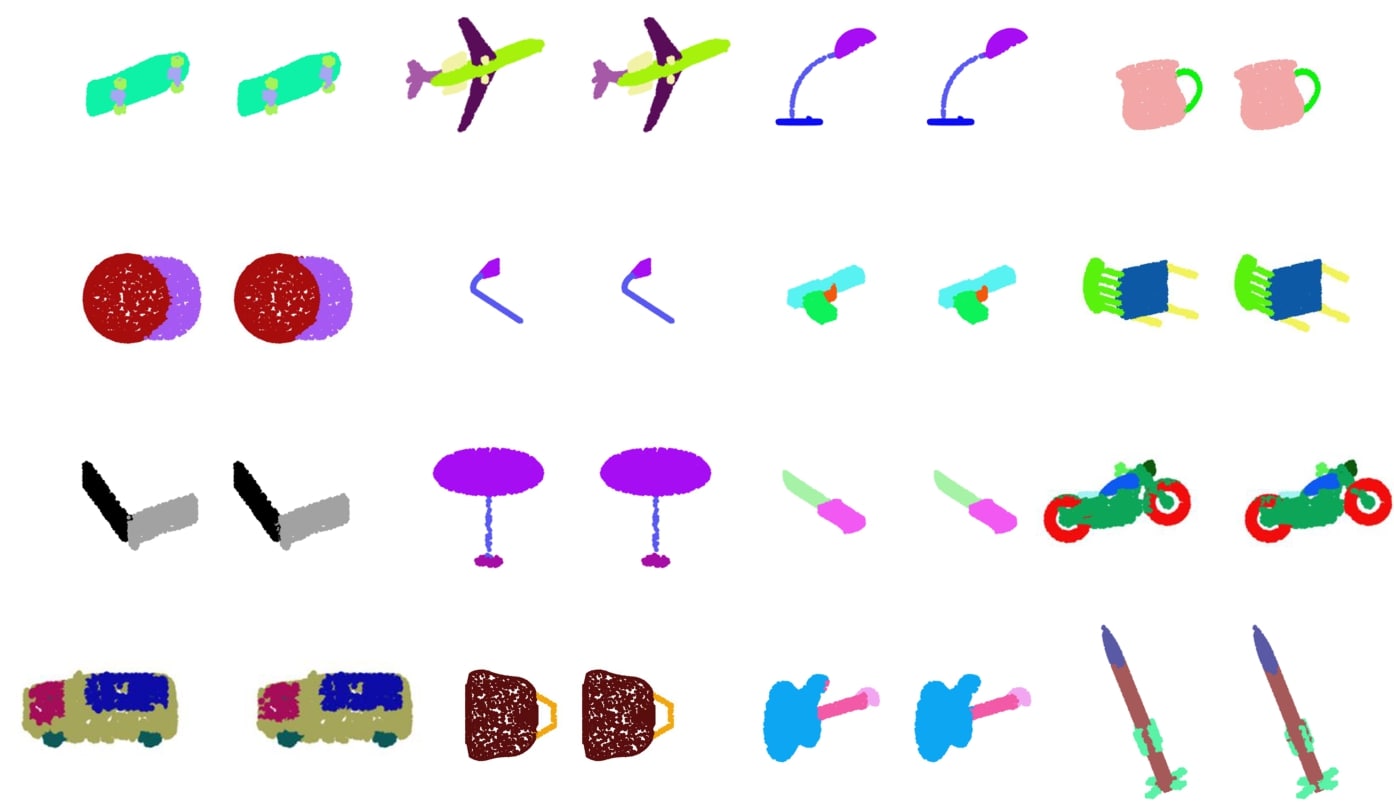}
\caption{Segmentation results on ShapeNet. First shape of each category is selected, where the left shape stands for the ground truth and the right one for our Pattern-Net.}
\label{fig::seg_resultsdd}
\end{figure}%#######################################

\textbf{Segmentation of given point clouds}: Segmentation of point cloud data is one of the popular 3D tasks. We carried out this experiment on the ShapeNet benchmark \cite{yi2016scalable} and followed the data split in \cite{qi2017pointnet}. ShapeNet contains 16881 models of 16 categories and they are labelled in 50 parts in total. Like \cite{yi2016scalable}, the Intersection-over-Union (IoU) of a shape is computed by averaging the IoUs of different parts in that shape, and the IoU of each category is obtained by averaging the IoUs of all the shapes belonging to that category. The results are summarized in Table \ref{table::seg_res}. When the training set is small, the performance of our network is on par with the other methods. Our technique extracts common deep features between shapes so it needs a sufficiently large number of training samples. From the table, it can be seen that Pattern-Net outperforms the existing methods on relatively large categories like airplane, car, chair and lamp. A sample result of each category is illustrated in Fig. \ref{fig::seg_resultsdd}.\\

\setlength{\tabcolsep}{1.3pt}%################################
\begin{table*} [ht]
\caption{Complexity of different methods for point cloud classification task} \label{table:noise} 
\centering 
\begin{tabular}{l cc cc cc cc cc c} 
\toprule % Top horizontal line
{Method} 
    & PointNet++
    & PointCNN & DGCNN
    & SO-Net & RS-CNN & DensePoint&Pattern-Net\\
\midrule 
\#params. &  1.48M &  8.2M & 11.8M & 11.5M & 1.41M & 670k&\bf{399k} \\ 
\bottomrule 
\label{table::complexity}
\end{tabular}
\end{table*}%################################
\textbf{Complexity analysis}:
Table \ref{table::complexity} reports the complexity of Pattern-Net as well as those of the existing techniques for the classification task. 
The number of input points was set to 1024. Thanks to the cloning decomposition, our technique needs less than 0.4M parameters which is much lower than  670k of DensePoint. 
This characteristic of our method is more appealing for real-time applications like mobile robotics and autonomous driving.

\section{Conclusions}
In this study, we have proposed a novel technique for efficiently learning stationary patterns in the given point clouds, which is less susceptible to noise, distortions and overfitting.
It is also invariant to changes in translation, rotation and scale.
The key idea is to decompose the point cloud into multiple subsets with similar structural information. Then we enforce the network to learn stable patterns.
Compared with noise that is order-less with unpredictable behaviour, natural objects have complex structures accompanied with irregularities in some parts due to the external disturbances.
Informative patterns could be successfully extracted if the level of randomness and uncertainty is diminished.
Unpredictable disorders cause inaccurate representation of given objects and this concept is known as `orderly disorder' theory.
To this end, we have proposed the cloning decomposing technique. Since our network learns just stable patterns, it is less prone to overfitting, which means it needs to train only once and can then run over a variety of data (e.g. different noise and sampling patterns). This method could provide a promising direction for robust representation of point cloud data.\\

\section*{Acknowledgements}
The authors gratefully acknowledge the HPC resources provided by Supercomputing Wales (SCW) and Aberystwyth University. M. Ghahremani acknowledges his AberDoc and President scholarships awarded by Aberystwyth University. Y. Liu and A. Behera are partially supported by BBSRC grant BB/R02118X/1 and UKIERI-DST grant CHARM (DST UKIERI-2018-19-10) respectively.
\bibliographystyle{splncs}
\bibliography{egbib}

\begin{thebibliography}{10}

\bibitem{maturana2015voxnet}
Daniel Maturana and Sebastian Scherer.
\newblock Voxnet: A 3d convolutional neural network for real-time object
  recognition.
\newblock In {\em 2015 IEEE/RSJ International Conference on Intelligent Robots
  and Systems (IROS)}, pages 922--928. IEEE, 2015.

\bibitem{wu20153d}
Zhirong Wu, Shuran Song, Aditya Khosla, Fisher Yu, Linguang Zhang, Xiaoou Tang,
  and Jianxiong Xiao.
\newblock 3d shapenets: A deep representation for volumetric shapes.
\newblock In {\em Proceedings of the IEEE conference on computer vision and
  pattern recognition}, pages 1912--1920, 2015.

\bibitem{su2015multi}
Hang Su, Subhransu Maji, Evangelos Kalogerakis, and Erik Learned-Miller.
\newblock Multi-view convolutional neural networks for 3d shape recognition.
\newblock In {\em Proceedings of the IEEE international conference on computer
  vision}, pages 945--953, 2015.

\bibitem{qi2016volumetric}
Charles~R Qi, Hao Su, Matthias Nie{\ss}ner, Angela Dai, Mengyuan Yan, and
  Leonidas~J Guibas.
\newblock Volumetric and multi-view cnns for object classification on 3d data.
\newblock In {\em Proceedings of the IEEE conference on computer vision and
  pattern recognition}, pages 5648--5656, 2016.

\bibitem{qi2017pointnet}
Charles~R Qi, Hao Su, Kaichun Mo, and Leonidas~J Guibas.
\newblock Pointnet: Deep learning on point sets for 3d classification and
  segmentation.
\newblock In {\em Proceedings of the IEEE Conference on Computer Vision and
  Pattern Recognition}, pages 652--660, 2017.

\bibitem{guerrero2018pcpnet}
Paul Guerrero, Yanir Kleiman, Maks Ovsjanikov, and Niloy~J Mitra.
\newblock Pcpnet learning local shape properties from raw point clouds.
\newblock {\em Computer Graphics Forum}, 37(2):75--85, 2018.

\bibitem{zhi2017lightnet}
Shuaifeng Zhi, Yongxiang Liu, Xiang Li, and Yulan Guo.
\newblock Lightnet: A lightweight 3d convolutional neural network for real-time
  3d object recognition.
\newblock In {\em 3DOR}, 2017.

\bibitem{ma2017bv}
Chao Ma, Wei An, Yinjie Lei, and Yulan Guo.
\newblock Bv-cnns: Binary volumetric convolutional networks for 3d object
  recognition.
\newblock In {\em BMVC}, page~4, 2017.

\bibitem{liu2019relation}
Yongcheng Liu, Bin Fan, Shiming Xiang, and Chunhong Pan.
\newblock Relation-shape convolutional neural network for point cloud analysis.
\newblock In {\em Proceedings of the IEEE Conference on Computer Vision and
  Pattern Recognition}, pages 8895--8904, 2019.

\bibitem{yi2016scalable}
Li~Yi, Vladimir~G Kim, Duygu Ceylan, I-Chao Shen, Mengyan Yan, Hao Su, Cewu Lu,
  Qixing Huang, Alla Sheffer, and Leonidas Guibas.
\newblock A scalable active framework for region annotation in 3d shape
  collections.
\newblock {\em ACM Transactions on Graphics (TOG)}, 35(6):1--12, 2016.

\bibitem{wang2019dominant}
Chu Wang, Marcello Pelillo, and Kaleem Siddiqi.
\newblock Dominant set clustering and pooling for multi-view 3d object
  recognition.
\newblock {\em arXiv preprint arXiv:1906.01592}, 2019.

\bibitem{riegler2017octnet}
Gernot Riegler, Ali Osman~Ulusoy, and Andreas Geiger.
\newblock Octnet: Learning deep 3d representations at high resolutions.
\newblock In {\em Proceedings of the IEEE Conference on Computer Vision and
  Pattern Recognition}, pages 3577--3586, 2017.

\bibitem{tatarchenko2017octree}
Maxim Tatarchenko, Alexey Dosovitskiy, and Thomas Brox.
\newblock Octree generating networks: Efficient convolutional architectures for
  high-resolution 3d outputs.
\newblock In {\em Proceedings of the IEEE International Conference on Computer
  Vision}, pages 2088--2096, 2017.

\bibitem{klokov2017escape}
Roman Klokov and Victor Lempitsky.
\newblock Escape from cells: Deep kd-networks for the recognition of 3d point
  cloud models.
\newblock In {\em Proceedings of the IEEE International Conference on Computer
  Vision}, pages 863--872, 2017.

\bibitem{qi2017pointnet++}
Charles~Ruizhongtai Qi, Li~Yi, Hao Su, and Leonidas~J Guibas.
\newblock Pointnet++: Deep hierarchical feature learning on point sets in a
  metric space.
\newblock In {\em Advances in neural information processing systems}, pages
  5099--5108, 2017.

\bibitem{huang2017densely}
Gao Huang, Zhuang Liu, Laurens Van Der~Maaten, and Kilian~Q Weinberger.
\newblock Densely connected convolutional networks.
\newblock In {\em Proceedings of the IEEE conference on computer vision and
  pattern recognition}, pages 4700--4708, 2017.

\bibitem{liu2019densepoint}
Yongcheng Liu, Bin Fan, Gaofeng Meng, Jiwen Lu, Shiming Xiang, and Chunhong
  Pan.
\newblock Densepoint: Learning densely contextual representation for efficient
  point cloud processing.
\newblock In {\em Proceedings of the IEEE International Conference on Computer
  Vision}, pages 5239--5248, 2019.

\bibitem{shen2018mining}
Yiru Shen, Chen Feng, Yaoqing Yang, and Dong Tian.
\newblock Mining point cloud local structures by kernel correlation and graph
  pooling.
\newblock In {\em Proceedings of the IEEE conference on computer vision and
  pattern recognition}, pages 4548--4557, 2018.

\bibitem{landrieu2018large}
Loic Landrieu and Martin Simonovsky.
\newblock Large-scale point cloud semantic segmentation with superpoint graphs.
\newblock In {\em Proceedings of the IEEE Conference on Computer Vision and
  Pattern Recognition}, pages 4558--4567, 2018.

\bibitem{wang2019dynamic}
Yue Wang, Yongbin Sun, Ziwei Liu, Sanjay~E Sarma, Michael~M Bronstein, and
  Justin~M Solomon.
\newblock Dynamic graph cnn for learning on point clouds.
\newblock {\em ACM Transactions on Graphics (TOG)}, 38(5):146, 2019.

\bibitem{xu2019grid}
Qiangeng Xu.
\newblock Grid-gcn for fast and scalable point cloud learning.
\newblock {\em arXiv preprint arXiv:1912.02984}, 2019.

\bibitem{zhang2019rotation}
Zhiyuan Zhang, Binh-Son Hua, David~W Rosen, and Sai-Kit Yeung.
\newblock Rotation invariant convolutions for 3d point clouds deep learning.
\newblock In {\em 2019 International Conference on 3D Vision (3DV)}, pages
  204--213. IEEE, 2019.

\bibitem{atzmon2018point}
Matan Atzmon, Haggai Maron, and Yaron Lipman.
\newblock Point convolutional neural networks by extension operators.
\newblock {\em arXiv preprint arXiv:1803.10091}, 2018.

\bibitem{peters1996chaos}
Edgar~E Peters.
\newblock {\em Chaos and order in the capital markets: a new view of cycles,
  prices, and market volatility}.
\newblock John Wiley \& Sons, 1996.

\bibitem{davood2015orderly}
SS~Davood.
\newblock Orderly disorder in modern physics.
\newblock {\em International Letters of Chemistry, Physics and Astronomy},
  48:163--172, 2015.

\bibitem{li2018so}
Jiaxin Li, Ben~M Chen, and Gim Hee~Lee.
\newblock So-net: Self-organizing network for point cloud analysis.
\newblock In {\em Proceedings of the IEEE conference on computer vision and
  pattern recognition}, pages 9397--9406, 2018.

\bibitem{simonovsky2017dynamic}
Martin Simonovsky and Nikos Komodakis.
\newblock Dynamic edge-conditioned filters in convolutional neural networks on
  graphs.
\newblock In {\em Proceedings of the IEEE conference on computer vision and
  pattern recognition}, pages 3693--3702, 2017.

\bibitem{brake2019singular}
Danielle~A Brake, Jonathan~D Hauenstein, Frank-Olaf Schreyer, Andrew~J Sommese,
  and Michael~E Stillman.
\newblock Singular value decomposition of complexes.
\newblock {\em SIAM Journal on Applied Algebra and Geometry}, 3(3):507--522,
  2019.

\end{thebibliography}
\end{document}